\documentclass[conference,a4paper]{IEEEtran}
\IEEEoverridecommandlockouts
\usepackage{cite}
\usepackage{amsmath,amssymb,amsfonts}
\usepackage{algorithmic}
\usepackage{graphicx}
\usepackage{textcomp}
\usepackage{xcolor}
\usepackage{chemformula}
\def\BibTeX{{\rm B\kern-.05em{\sc i\kern-.025em b}\kern-.08em
    T\kern-.1667em\lower.7ex\hbox{E}\kern-.125emX}}
\begin{document}

\title{Neural Network Surrogates for Free Energy Computation of Complex Chemical Systems\\
}


\author{\IEEEauthorblockN{Wasut Pornpatcharapong}
\IEEEauthorblockA{\textit{International College of Digital Innovation} \\
\textit{Chiang Mai University}\\
Chiang Mai, Thailand \\
wasut.p@cmu.ac.th 
}\thanks{979-8-3315-7383-6/25/\$31.00 © 2025 IEEE}
}

\maketitle

\begin{abstract}
Free energy reconstruction methods such as Gaussian Process Regression (GPR) require Jacobians of the collective variables (CVs), a bottleneck that restricts the use of complex or machine-learned CVs. We introduce a neural network surrogate framework that learns CVs directly from Cartesian coordinates and uses automatic differentiation to provide Jacobians, bypassing analytical forms. On an \ch{MgCl2} ion-pairing system, our method achieved high accuracy for both a simple distance CV and a complex coordination-number CV. Moreover, Jacobian errors also followed a near-Gaussian distribution, making them suitable for GPR pipelines. This framework enables gradient-based free energy methods to incorporate complex and machine-learned CVs, broadening the scope of biochemistry and materials simulations.

\end{abstract}


\section{Introduction}

High-dimensional collective variables (CVs) are essential for mapping free energy landscapes and extracting reaction mechanisms such as minimum free energy 
paths (MFEPs) \cite{E2002}. These landscapes underlie key processes in catalysis, ion transport, and biomolecular function \cite{Pornpatcharapong2025}, but 
are difficult to compute due to rare events and high dimensionality. Machine learning surrogates such as Gaussian Process Regression (GPR) and neural networks 
have accelerated free energy surface construction \cite{Mones2016, Sidky2018, P-ChemRev-2021-Deringer}. However, these methods require analytical Jacobians of CVs, 
which are infeasible for complex descriptors or machine-learned CVs, creating a critical bottleneck. 

To address this critical limitation, the autograd feature of a neural network (NN) surrogate framework was introduced to learn the Jacobians, completely 
bypassing analytical calculations. This work serves as a foundational proof-of-concept where the autograd-derived Jacobians are accurrate and computationally 
efficient relative to analytical computations for both simple and complex CVs. We rigorously test this on the \ch{MgCl2} ion-pairing system, proving that this 
method is a viable, scalable, and essential first step towards incorporating complex, machine-learned CVs into GPR-based free energy calculations.

\section{Related Work}

\subsection{Theoretical Background}

Under the canonical ensemble, where a chemical system has constant number of particles ($N$), volume ($V$) and temperature ($T$), 
the free energy of a system ($A$) is defined in the Cartesian space as \eqref{eq:fe-defn} \cite{mcquarrie76a},

\begin{equation}\begin{aligned}
    A(\mathbf{r}) &= -\beta^{-1} \ln Q(\mathbf{r}) \\
                  &= -\beta^{-1} \ln\int e^{-\beta V(\mathbf{r})} d\mathbf{r}  
 \label{eq:fe-defn}
\end{aligned}\end{equation}

\noindent where $\beta = 1/k_B T$, $k_B$ is the Boltzmann’s constant, and $\mathbf{r}$ are 
Cartesian coordinate vectors. However, the free energy landscape is usually represented by surrogate features representing 
chemically relevant portions of the system, termed \textit{collective variables} (CV; $\xi$), which itself is a function of all 
Cartesian coordinates $\xi(\mathbf{r})$. When projected onto the CV space, free energy landscape can be 
written as \eqref{eq:fe-cv}

\begin{equation}
    \begin{aligned}
      A(\boldsymbol{\xi}^*) &= -\beta^{-1} \ln \mathbf{P}(\boldsymbol{\xi}^*) \\
        &= -\beta^{-1} \ln\int \delta\left(\boldsymbol{\xi}(\mathbf{r}) - \boldsymbol{\xi}^*\right) 
        \,e^{-\beta V(\mathbf{r})} d\mathbf{r}
      \label{eq:fe-cv} 
    \end{aligned}
\end{equation}

\noindent where $\boldsymbol{\xi^*}$ are instantaneous positions of the CVs and $\mathbf{P}(\boldsymbol{\xi}^*)$ is the probability of finding 
the system at that particular position \cite{Mones2016}. Equations \eqref{eq:fe-defn} and \eqref{eq:fe-cv} imply that 
$\mathbf{P}(\boldsymbol{\xi}^*) \propto e^{-\beta A(\boldsymbol{\xi}^*)}$, which means the probability of finding a particular set of 
CV at high free energy tends to zero, and the high free energy region usually indicates the barrier between the reactant and the product 
of a chemical reaction. Thus, physical simulations alone cannot thoroughly sample this rare event region, yet crucial in understanding the 
entire reaction process, in a meaningful simulation timescale.

Enhanced sampling methods like Metadynamics (MTD) and its well-tempered variant (WT-MTD) \cite{Laio2002, Barducci2010} overcome this by adding a history dependent bias potential
defined in \eqref{eq:mtd}, but do not directly yield the free energy gradients. Therefore, simulation trajectories from such
methods need to be systematically unbiased to obtain the true free energy gradients needed for robust reconstruction methods like GPR.

\begin{equation}
V_\mathrm{MTD}(\boldsymbol{\xi}^*,t) = \sum_{t' < t} w(t') \exp\Bigg[ -\frac{(\boldsymbol{\xi}^* - \boldsymbol{\xi}(t'))^2}{2 \sigma^2} \Bigg]
\label{eq:mtd}
\end{equation}

\subsection{Machine Learning and Free Energy Computation}

Mones et al. proposed a framework that used GPR to reconstruct a smooth free energy landscape from the training data of WT-MTD simulations \cite{Mones2016}. In this framework, the ability of GPR to learn the function from its derivative is exploited, where the input for free energy reconstruction is called the unbiased instantaneous collective force (ICF; $\mathbf{f}$), which is simply the negative of the free energy gradient. $\mathbf{f}$, which is a matrix with number of columns equal to the number of CVs, can be calculated from \eqref{eq:GPR-ICF}, 

\begin{equation}
    \mathbf{f} = \frac{d}{dt}\left(\mathbf{Z}^{-1}\frac{d\boldsymbol{\xi}}{dt}\right)-\mathbf{F}_{\text{biased}}(\boldsymbol{\xi}) 
\label{eq:GPR-ICF}
\end{equation}

\noindent where $\mathbf{F}_{\text{biased}}(\boldsymbol{\xi})$ represents the negative value of the gradient of MTD potential as noted in \eqref{eq:mtd}, and $\mathbf{Z}$ is the metric tensor,

\begin{equation}
    \mathbf{Z} = \nabla_{\mathbf{r}}\boldsymbol{\xi}^T\mu^{-1}
    \nabla_{\mathbf{r}}\boldsymbol{\xi}
    \label{eq:GPR-tensor}
\end{equation}

\noindent where $\nabla_{\mathbf{r}}\boldsymbol{\xi}$ represents the Jacobian of the CVs with respect to the Cartesian coordinates of all atoms in the system in all dimensions, while $\mu$ is a diagonal matrix of atomic mass.

\subsection{Neural Networks and Autograd for Jacobian Computation}

Despite the potential to revolutionize free energy computations, the key limitation of Mones et al.’s framework is the difficulty of 
Jacobian computation for complex CVs. As demonstrated in \eqref{eq:GPR-tensor}, the framework requires Jacobian computation with respect 
to the Cartesian coordinates of all atoms in the system. Recent work by Pornpatcharapong extended Mones et al.’s work by relying 
on analytical evaluation of the Jacobians \cite{P-Molecules-2025-Wasut}. While analytical evaluation of the Jacobians is valid for 
simple CVs with analytical forms, in complex chemical systems, such as reactions in large biomolecules or novel material's superstructures, 
CVs may have complex or non-analytical forms or may even be derived 
from machine learning itself, rendering analytical evaluation impossible.

The integration of machine learning into computational science has given rise to the burgeoning field of Scientific Machine Learning 
(SciML) \cite{Baker2019}. A cornerstone of SciML is the use of automatic differentiation (autograd), a technique enabled by modern deep 
learning frameworks like PyTorch \cite{paszke2019pytorchimperativestylehighperformance} and TensorFlow \cite{tensorflow2015-whitepaper}, which allows for the efficient and accurate computation of derivatives \cite{JMLR:v18:17-468}. 
This capability is not merely a convenience for training networks but a powerful tool for solving scientific problems. For instance, 
Physics-Informed Neural Networks (PINNs) leverage autograd to solve partial differential equations by embedding physical laws directly 
into the loss function \cite{Raissi2019}. In these applications, the NN acts as a differentiable surrogate model, learning a function 
from data while providing immediate access to its derivatives through the autograd engine.

In the context of the potential application for this work, the power of autograd has been harnessed for discovery of non-linear 
CVs, targeting the most chemically relevant parts of any chemical systems \cite{Wehmeyer2018}. These machine-learned CVs, however, 
do not have 
analytical formulae. While autograd was also mentioned as a viable tool for Jacobian-based simulations \cite{ford2022automatic}, the specific application of using a NN surrogate to approximate pre-defined, 
complex collective variables for the sole purpose of calculating their Jacobians - to overcome the bottleneck in GPR for free energy reconstruction - remains unexplored. This represents a critical gap, as it is the key to unlocking the use 
of highly expressive, chemically relevant CVs in robust free energy calculations.

In this work, a neural network framework that acts as a surrogate for any CV function is proposed, leveraging autograd to provide 
accurate Jacobian estimates, thereby overcoming the primary bottleneck preventing the use of complex CVs in GPR-based free 
energy reconstruction. This work aims to serve as a proof-of-concept of applying CV Jacobian estimates from autograd of neural 
networks-learned CVs from Cartesian coordinates, employing both analytically simple and complex CVs for proper validation against 
analytical solutions.

\section{Methodology}

\subsection{Simulation Setup and CV Definitions}\label{subsec:meth-setup}

To assess the feasibility of future applications of the NN-based Jacobian computation in the realm of free energy computation for complex chemical systems, the simulation setup is based on the recent work by Pornpatcharapong, where the number of water molecules and ions, the physical constraints of the system, simulation time, and simulation parameters, including the WT-MTD bias, were directly modeled \cite{P-Molecules-2025-Wasut}. The only difference in this work with respect to Pornpatcharapong’s is the data collection rate, where this work saved the molecular dynamics trajectory at the rate of 1 step per frame, whereas the rate is 1,000 steps per frame in Pornpatcharapong’s work.

As this work sought to model NN-based Jacobian computation with both mathematically simple and complex CVs, namely the distance between the \ch{Mg^{2+}} ion and a chosen \ch{Cl^-} ion ($d$), which is simply the Euclidean distance between them as defined in \eqref{eq:cv-dmgcl},

\begin{equation}
    d = \sqrt{(x_{\mathrm{Mg}} - x_{\mathrm{Cl}})^2 + (y_{\mathrm{Mg}} - y_{\mathrm{Cl}})^2 + (z_{\mathrm{Mg}} - z_{\mathrm{Cl}})^2}
    \label{eq:cv-dmgcl}
\end{equation}

\noindent which necessitates 6-dimensional input vectors for this CV, namely $[x_{\mathrm{Mg}}, y_{\mathrm{Mg}}, z_{\mathrm{Mg}}, x_{\mathrm{Cl}}, y_{\mathrm{Cl}}, z_{\mathrm{Cl}}]$. On the other hand, the first hydration shell around the \ch{Mg^{2+}} ion ($C$), is both a chemically relevant CV for this system while also analytically complex. $C$ is defined as follow in \eqref{eq:cv-cn}

\begin{equation}
    C = \sum_{i=1}^{N} \frac{1}{2} \left\{ 1 + \tanh\left[ k \cdot (r_0 - d_{\mathrm{Mg-O}_i}) \right]\right\}
    \label{eq:cv-cn}
\end{equation}

\noindent where $r_0 = 0.265\,\mathrm{nm}$ and $k = 30\,\mathrm{nm}^{-1}$. $d_{\mathrm{Mg-O}_i}$ is the distance between the \ch{Mg^{2+}} ion and the $i$-th oxygen atom of a water molecule. The value of $C$ is usually computed using all water molecules in the system (629 in this case). Nevertheless, as this system chemically can never have the value of $C$ greater than 10, the coordinates of 20 oxygen atoms in their respective water molecules were used as the input. Along with the coordinate of the \ch{Mg^{2+}} ion, the input vectors for this CV have 63 dimensions. Therefore, the input vector structure for $C$ is $[x_{\mathrm{Mg}}, y_{\mathrm{Mg}}, z_{\mathrm{Mg}}, x_{\mathrm{O}_1}, y_{\mathrm{O}_1}, \ldots, z_{\mathrm{O}_{20}}]$.

\subsection{Data Generation}

To rigorously evaluate the proposed framework, two distinct datasets of 50,000 samples each were constructed, after which PyTorch was used
to build the NN models of all datasets, 80\% of which was used for training, and the rest as test.

\begin{enumerate}
\item A simulation dataset derived from the molecular dynamics trajectory.
\item A randomized benchmark dataset with coordinates uniformly sampled within the simulation box, serving as a control case.
\end{enumerate}

The simulation dataset was designed to capture the chemically relevant region of the free energy landscape, including the rare event of ion pairing and solvent exchange. A contiguous segment was selected from the trajectory to ensure coverage of this transition, where 50,000 frames were systematically selected at a uniform interval to create a manageable yet representative dataset that preserves the temporal progression and physical distribution of states.

The randomized dataset was generated by sampling atomic coordinates from a uniform distribution across the volume of the simulation box. This dataset provides a stress test for the neural network, challenging it to learn the CV functions across the entire configurational space, including high-energy, physically unrealistic regions that are poorly sampled in standard simulations. The simulation and subsequent NN modeling were performed on a local workstation with an 8-core AMD Ryzen\textsuperscript{TM} 7 5800X CPU with 32GB of DDR4 RAM and NVidia GeForce\textsuperscript{TM} RTX 2060 GPU.

\subsection{Neural Network Architecture and Training}

The core of this approach is a feed-forward NN (a multi-layer perceptron, or MLP) designed to function as a differentiable surrogate model for CVs. The network learns the mapping from atomic configurations to CV values, and its automatic differentiation provides the necessary Jacobians. The input to the network is a vector of Cartesian coordinates.

A critical feature of our architecture is the hard-coding of physical constraints. Before the first layer, input coordinates are transformed to enforce periodic boundary conditions (PBC) using the minimum image convention, ensuring the model only operates on coordinates within the primary simulation box of length $L = 2.7\,\mathrm{nm}$,

\begin{equation}
    \mathbf{r}_{\text{normalized}} = \text{remainder}(\mathbf{r} + L/2, L) - L/2
\end{equation}

While periodic boundary conditions have been implemented in prior machine learning contexts, such as neural networks with periodic layers \cite{dong2021method} or PINNs with boundary-enforcing losses \cite{straub2025hard}, to our knowledge, this is the first time that PBCs are directly encoded into a neural network surrogate for providing CV values and Jacobians in GPR-based free-energy reconstruction. This transformation guarantees that the input to the network is continuous across periodic boundaries, a necessary condition for learning a smooth function.

The network architecture consists of an input layer, multiple hidden layers with ReLU (Rectified Linear Unit) activation functions, and a linear output layer. The width of the hidden layers was set to $[64, 128, 64, 32]$, providing sufficient non-linear capacity to learn the complex mapping from coordinates to CV values. Dropout layers with a rate of $p = 0.1$ were included between hidden layers to regularize the model and prevent overfitting.

The network was trained to minimize the Mean Squared Error (MSE) between its predictions $\hat{y}_i$ and the ground-truth CV values $y_i$ calculated analytically for each data point $i$ in a batch of size $N$:

\begin{equation}
    \mathcal{L} = \frac{1}{N} \sum_{i=1}^{N} (y_i - \hat{y}_i)^2
\end{equation}

The model parameters (weights $\mathbf{W}$ and biases $\mathbf{b}$) were optimized using the Adam optimizer with a learning rate of $\alpha = 10^{-3}$ and L2 weight decay of $\lambda = 10^{-5}$. A \texttt{ReduceLROnPlateau} scheduler was employed to reduce the learning rate by a factor of 0.5 upon plateau of the validation loss, facilitating finer convergence.

The output layer produces a single, scalar value $\hat{y}$ representing the predicted CV. For the distance CV, an absolute value activation $|\hat{y}|$ was applied to ensure a physically plausible non-negative output. For the coordination number CV, no output activation was needed as the network successfully learned the bounds of the switching function defined in equation ~\eqref{eq:cv-cn}. Numerical stability for the complex $C$ output was ensured within the custom loss function by clamping the input to the rational function and adding a small epsilon ($\epsilon = 10^{-8}$) to denominators to prevent division by zero during the training process.

After training, the Jacobian of the learned function, the partial derivatives of the output CV with respect to each input coordinate, is computed automatically using PyTorch's automatic differentiation (autograd) engine. For a given input vector $\mathbf{x}$, the Jacobian $\mathbf{J}$ is given by:

\begin{equation}
    \mathbf{J} = \nabla_{\mathbf{x}} \hat{y} = \left[ \frac{\partial \hat{y}}{\partial x_1}, \frac{\partial \hat{y}}{\partial x_2}, \ldots, \frac{\partial \hat{y}}{\partial x_D} \right]
\end{equation}

\noindent where $D$ is the dimension of the input. This is achieved efficiently via a backward pass through the computational graph, calculating the gradient of the output with respect to the inputs.

\section{Results}

\subsection{Prediction of $d$ and Its Jacobians}

\begin{figure}[htbp]
\centerline{\includegraphics[width=0.95\linewidth]{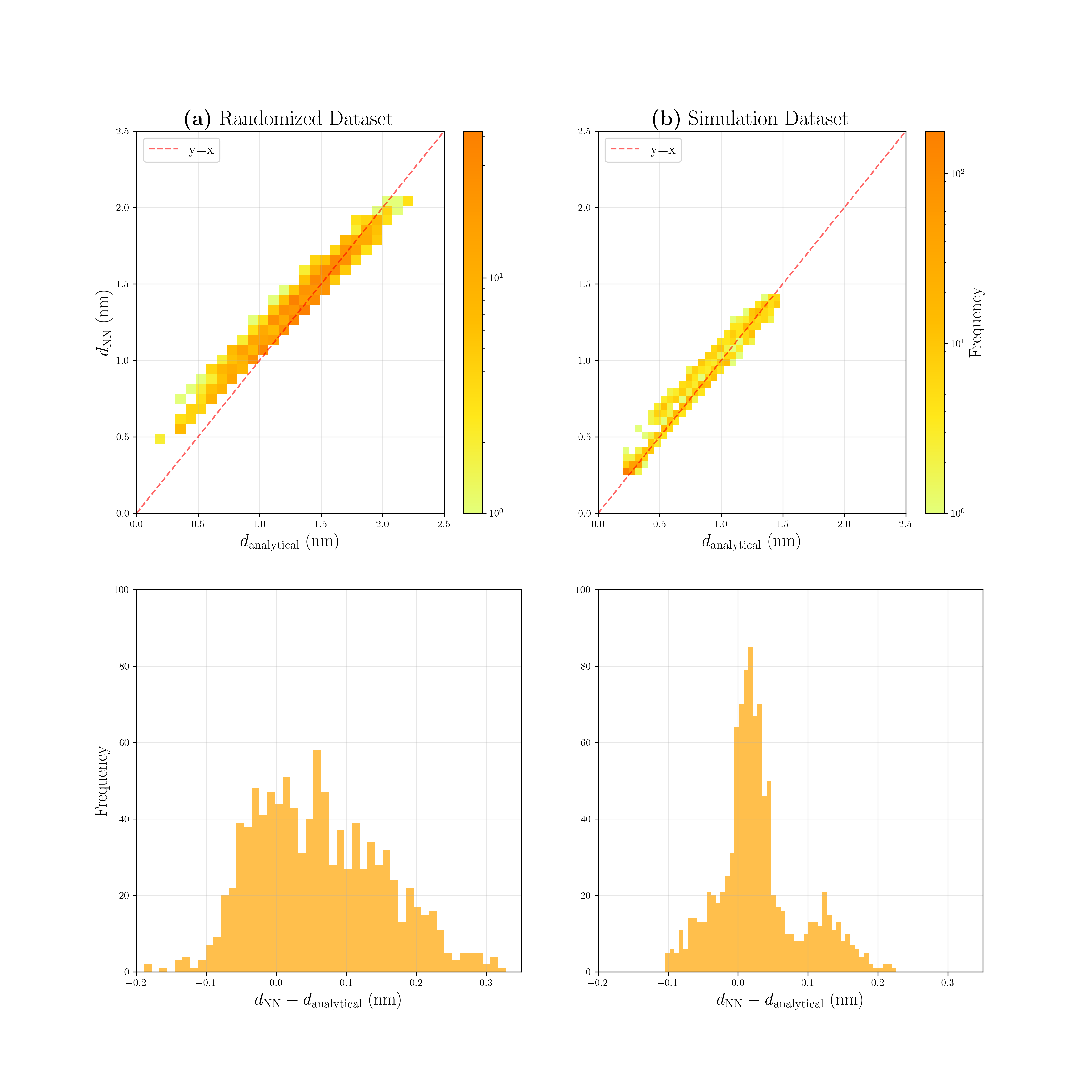}}
\caption{Prediction of $d$ for the randomized dataset \textbf{(a)} and simulation dataset \textbf{(b)} and distribution of prediction errors.}
\label{fig:distance}
\end{figure}

The prediction accuracy for the CV value itself was superior on the simulation dataset. The model achieved a root mean squared error (RMSE) of 0.066 nm 
and a mean absolute error (MAE) of 0.047 nm on the simulation data, outperforming its performance on the randomized data (RMSE: 0.111 nm, MAE: 0.086 nm) 
(Table \ref{tab}). The range of predicted distances also differed significantly between the two datasets. Predictions on the randomized dataset spanned 
from 0.2 to 2.3 nm (Fig. \ref{fig:distance}), while predictions on the simulation dataset were confined to a range of 0.2 to 1.5 nm (Fig. \ref{fig:distance}), 
a consequence of the physical restraints inherent to the simulation. This bimodal error distribution is not a model failure but a signature of the underlying 
physics, corresponding to the two dominant ion-pair configurations (contact and solvent-separated), a finding consistent with previous work \cite{P-Molecules-2025-Wasut}.

The distribution of prediction errors further highlights this divergence (Fig. \ref{fig:distance}). While both distributions show a primary peak near zero 
error, confirming general agreement with the analytical calculation (Eq. \eqref{eq:cv-dmgcl}), the error distribution for the simulation data features a second peak.

\begin{figure}[htbp]
\centerline{\includegraphics[width=0.95\linewidth]{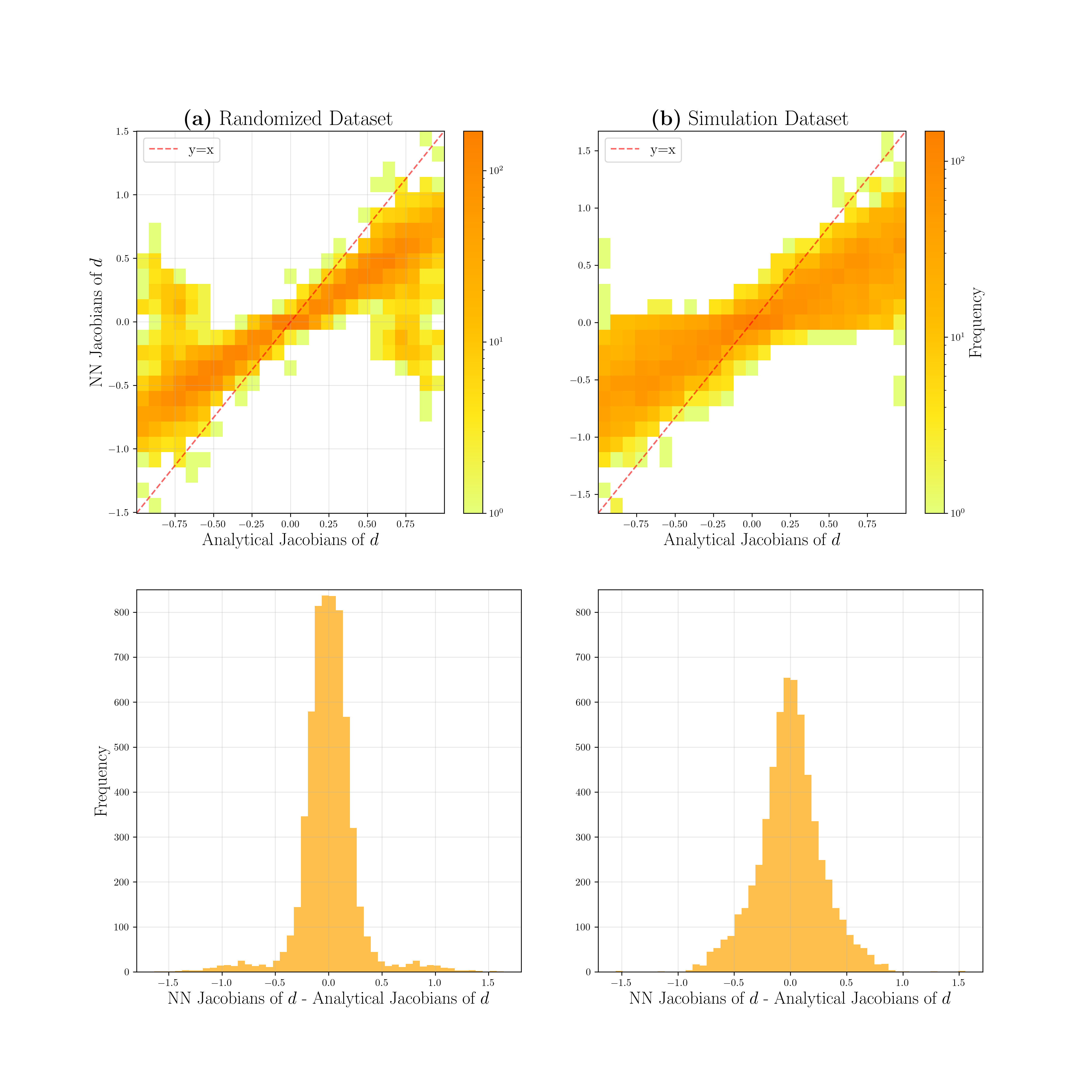}}
\caption{Prediction of the Jacobians (all dimensions) of $d$ for the randomized dataset \textbf{(a)} and simulation dataset \textbf{(b)} and distribution of prediction errors.}
\label{fig:distance_j}
\end{figure}

For the Jacobian of the $d$, the model demonstrated high accuracy on both datasets, as evidenced by the concentration of data points along the line of 
unity in the heatmaps (Fig. \ref{fig:distance_j}). However, quantitative metrics indicate a subtle difference: performance was slightly better on the 
randomized data (Jacobian MAE: 0.171) than on the simulation data (Jacobian MAE: 0.215) (Table \ref{tab}). This is visually corroborated by a broader spread 
of points deviating from the $y=x$ line in the simulation data heatmap. The distribution of prediction errors for the Jacobians (Fig. \ref{fig:distance_j}) 
also showed a unimodal, near-Gaussian distribution centered at zero. While a formal normality test is beyond this scope, this distribution shape is 
highly desirable for GPR pipelines, which are adept at modeling such well-behaved, zero-mean noise. 

\begin{table}[htbp]
\caption{Aggregated Prediction Errors for Randomized and Simulation Datasets for Both CVs.}
\begin{center}
\begin{tabular}{|c|c|c|c|c|}
\hline
\textbf{Error} & \textbf{$d_{\mathrm{Random}}$} & \textbf{$d_{\mathrm{Simulation}}$} & \textbf{$C_{\mathrm{Random}}$}& \textbf{$C_{\mathrm{Simulation}}$} \\
\hline
RMSE & 0.111 nm & 0.066 nm & 0.277 & 0.129 \\
MAE & 0.086 nm & 0.047 nm & 0.103 & 0.037 \\
RMSE ($\mathbf{J}$) & 0.262 & 0.286 & 0.611 & 0.490 \\
MAE ($\mathbf{J}$) & 0.171 & 0.215 & 0.061 & 0.206 \\
\hline
\end{tabular}
\label{tab}
\end{center}
\end{table}

\subsection{Prediction of the $C$ and Its Jacobians}

The prediction of the $C$ with our network proved more challenging than for $d$, a difficulty attributed to the switching function's near-discontinuous 
nature in \eqref{eq:cv-cn}, creating a quasi-classification problem.

\begin{figure}[htbp]
\centerline{\includegraphics[width=0.95\linewidth]{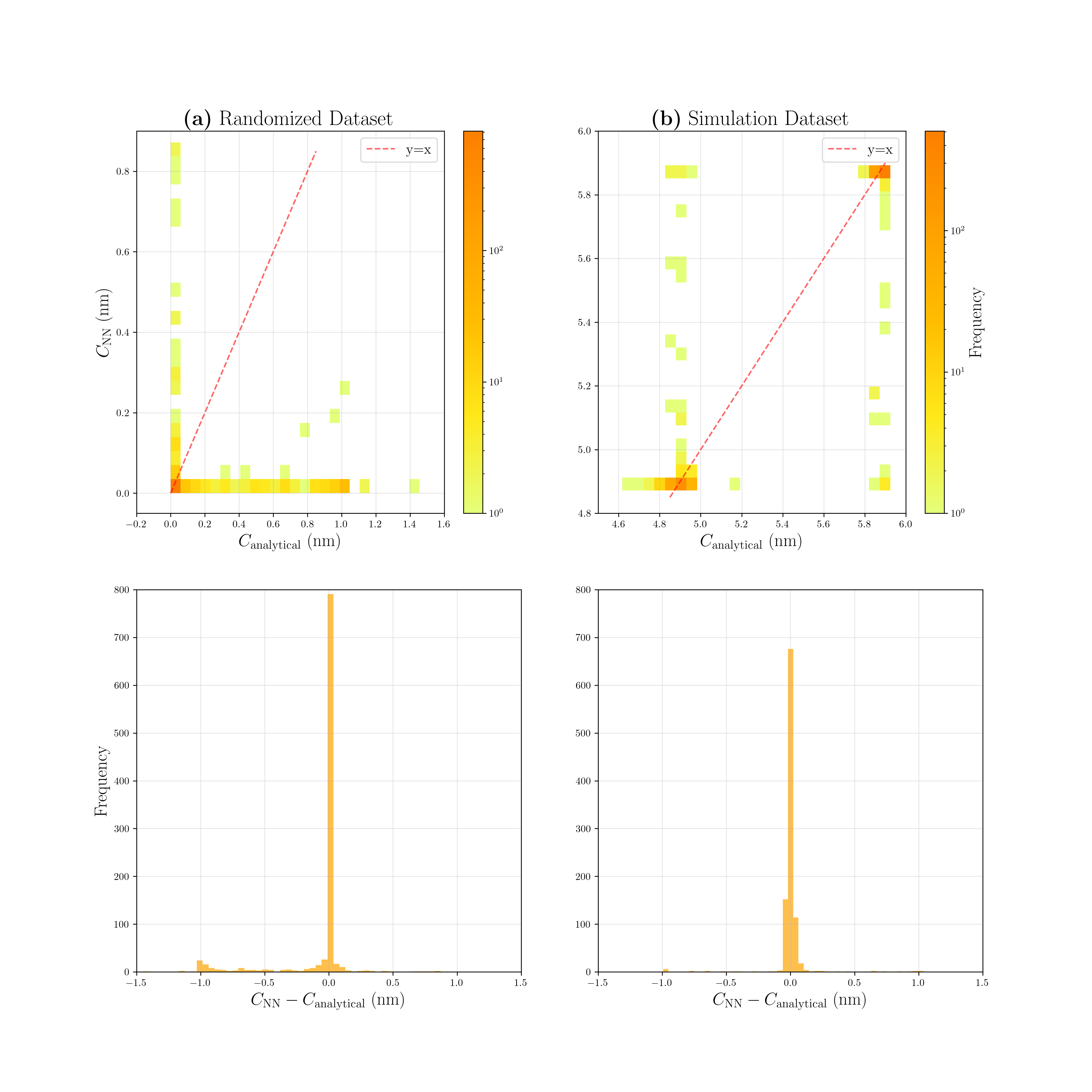}}
\caption{Prediction of $C$ for the randomized dataset \textbf{(a)} and simulation dataset \textbf{(b)} and distribution of prediction errors.}
\label{fig:cn}
\end{figure}

Despite this challenge, the network learned the values of $C$ effectively, demonstrating markedly superior performance on the simulation dataset (RMSE: 0.129, 
MAE: 0.037) compared to the randomized dataset (RMSE: 0.277, MAE: 0.103) (Table \ref{tab}). This result confirms that the network more accurately classifies 
the coordination number within physically realistic configurational regimes. The heatmaps of predicted vs. analytical $C$ values show general agreement with the 
unity line for both datasets, though significant deviations are present, reflecting the function's inherent complexity (Fig. \ref{fig:cn}).

\begin{figure}[htbp]
\centerline{\includegraphics[width=0.95\linewidth]{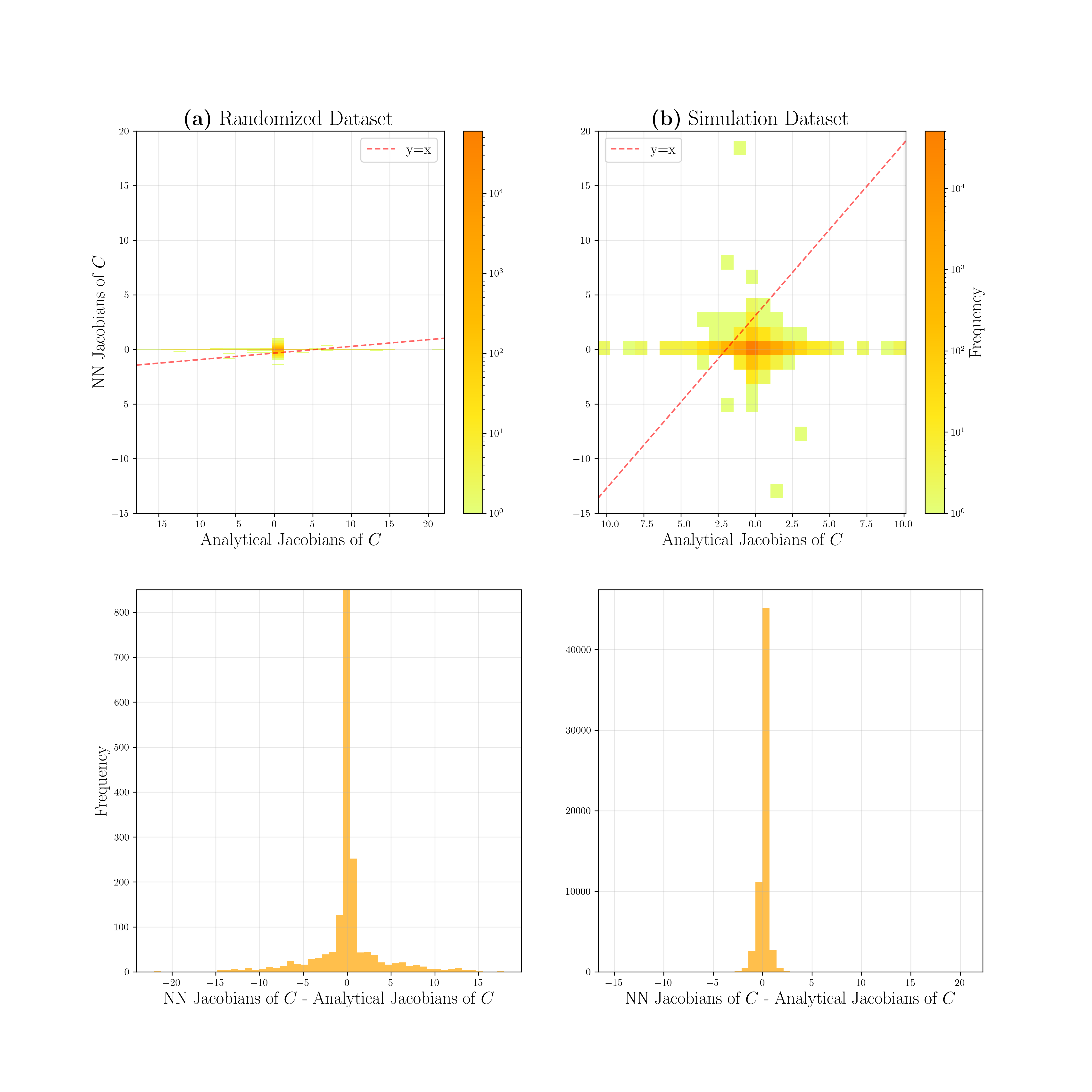}}
\caption{Prediction of the Jacobians (all dimensions) of $C$ for the randomized dataset \textbf{(a)} and simulation dataset \textbf{(b)} and distribution of prediction errors.}
\label{fig:cn_j}
\end{figure}

The prediction of the Jacobians for $C$ revealed a more pronounced performance dichotomy. 
Quantitative metrics showed large deviations from analytical values for both datasets (Table \ref{tab}). However, the heatmaps in Fig. \ref{fig:cn_j} 
reveal the fundamentally different nature of the error. The prediction of the Jacobians of $C$, when summing all dimensions, is largely zero even when the 
analytical Jacobians yielded a wide range of results for the randomized dataset (Fig. \ref{fig:cn_j}(a)). In contrast, the simulation dataset's predictions, while having a higher 
MAE (0.206), exhibit a more uniform error distribution across the range of analytical values (Fig. \ref{fig:cn_j}(b)). The model's poor performance on the 
randomized dataset for $C$ Jacobians (Fig. \ref{fig:cn_j}(a)) highlights a key finding: the network 
struggles with out of distribution, high gradient regions. However, its superior performance on the physically relevant simulation data (Fig. \ref{fig:cn_j}(b)), where 
the Jacobian error is tightly peaked at zero, demonstrates its ability to specialize and achieve high fidelity in the exact configurational space that matters 
for free energy calculations - the low energy basins where the system resides, a finding that reconciles the lower RMSE with the higher MAE and indicates a 
superior prediction for the thermodynamically relevant configurational space.

\section{Discussion}

The performance of our NN surrogate reveals critical insights into the learning process for the CVs. For the $d$ CV, the network successfully predicted values across the entire range permitted by the simulation's periodic boundary conditions (0 to 2.34 nm). More importantly, its performance on the simulation data captured a key chemical phenomenon. The bimodal distribution of prediction errors (Fig. \ref{fig:distance}) is not a shortcoming but a signature of the two dominant ion-pair configurations: the contact ion pair (CIP) and the solvent-separated ion pair (SSIP) \cite{P-Molecules-2025-Wasut}. This indicates that the network's error is structurally linked to the underlying physical states of the system. Furthermore, the near-Gaussian distribution of errors in the Jacobian prediction for $d$ suggests a high degree of numerical reliability.

The prediction of $C$ presented a distinct challenge due to the step-like nature of its switching function, effectively framing the task as a quasi-classification problem. The network's superior accuracy on the simulation data, evidenced by a tighter error distribution (Fig. \ref{fig:cn}), demonstrates its ability to specialize in thermodynamically relevant regions of configuration space. This specialization also explains the sharp, peaked distribution of Jacobian errors near zero for the simulation data. The model excels at predicting the derivatives precisely where the system spends most of its time (where the CV is flat and its derivative is zero), which is the most critical regime for accurately mapping the free energy basins.

The key advantage is independence from CV functional form. The network treats CVs as black boxes, learning values and Jacobians directly from data. Unlike the recent work on GPR \cite{P-Molecules-2025-Wasut}, which employed analytical Jacobians, the present NN surrogate bypasses this bottleneck, enabling non-analytical and machine-learned CVs to be incorporated into GPR pipelines. 

A critical insight from this study is that while Jacobian errors exist, their distribution for both CVs on the simulation data follows a Gaussian pattern. This is a highly desirable characteristic. The inherent ability of Gaussian Process Regression to model and filter out well-behaved, zero-mean noise means that the error profiles of our NN-predicted Jacobians are not only acceptable but are of a type that the subsequent free energy reconstruction pipeline is designed to handle effectively.  The architecture is also inherently scalable, which means the framework can be directly extended to high-dimensional input spaces, enabling the study of more complex systems such as biomolecular folding or catalytic reactions where the relevant collective variables involve many atoms. The computational efficiency of the method, especially once trained, offers a significant advantage over traditional approaches for these demanding applications.

\section{Conclusion}

The NN surrogate framework introduced in this work successfully bypasses critical Jacobian issues in free energy computations. By leveraging automatic 
differentiation, it provides a computationally efficient method that is agnostic to the analytical form of any CV. Our model yields accurate Jacobian 
estimates for both simple and complex CVs and respects the underlying physics of the simulation. The near-Gaussian error profile of these Jacobians makes 
them ideally suited for GPR pipelines, which inherently handle such noisy training data. This work serves as an essential proof-of-concept, unlocking a 
pathway to incorporate complex, machine-learned CVs into robust, multidimensional free energy landscape reconstructions. Future work will focus on a full 
GPR pipeline integration and applications to large-scale, chemically important systems.

\bibliographystyle{IEEEtran}
\bibliography{ICSEC2025_testbib}

\end{document}